\documentclass[a4paper]{article}
\usepackage[utf8]{inputenc}
\usepackage{INTERSPEECH2019}
\usepackage{url}
\usepackage{hyperref}
\usepackage{multirow}
\usepackage{cite}  %
\usepackage{color}

\usepackage{cleveref}

\usepackage{acro}

\DeclareAcronym{asr}{
        short = ASR ,
        long = automatic speech recognition ,
}

\DeclareAcronym{bpe}{
        short = BPE ,
        long = byte-pair encoding ,
}

\DeclareAcronym{cart}{
	short = CART ,
	long = classification and regression tree ,
}

\DeclareAcronym{cmllr}{
	short = CMLLR ,
	long = constrained maximum likelihood linear regression ,
}

\DeclareAcronym{dnn}{
	short = DNN ,
	long = deep neural network ,
}

\DeclareAcronym{gmm}{
	short = GMM ,
	long = gaussian mixture model ,
}

\DeclareAcronym{gmmhmm}{
	short = GMM/HMM ,
	long = Gaussian mixture model/hidden Markov model ,
}

\DeclareAcronym{hmm}{
	short = HMM ,
	long = hidden Markov model ,
}

\DeclareAcronym{lda}{
	short = LDA ,
	long = linear discriminant analysis ,
}

\DeclareAcronym{lfmmi}{
	short = LF-MMI ,
	long = lattice-free maximum mutual information ,
}

\DeclareAcronym{smbr}{
	short = sMBR ,
	long = state-level minimum bayes risk ,
}

\DeclareAcronym{lm}{
	short = LM ,
	long = language model ,
}

\DeclareAcronym{lstm}{
	short = LSTM ,
	long = long short-term memory ,
}

\DeclareAcronym{mfcc}{
	short = MFCC ,
	long =  Mel frequency cepstral coefficients ,
}

\DeclareAcronym{nn}{
	short = NN ,
	long = neural network ,
}

\DeclareAcronym{ocd}{
	short = OCD ,
	long = optimal completion distillation ,
}

\DeclareAcronym{sat}{
	short = SAT ,
	long =  speaker adaptive training ,
}

\DeclareAcronym{tdnn}{
	short = TDNN ,
	long = time delay neural network ,
}

\DeclareAcronym{vtln}{
	short = VTLN ,
	long =  vocal tract length normalization ,
}

\DeclareAcronym{wer}{
	short = WER ,
	long =  word error rate ,
}

\def\acknowledgements{\begin{center}
{\bf \large Acknowledgements\vspace{-.5em}\vspace{2pt}}
\end{center}}

\def\thebibliography#1{\begin{center}
{\bf \large References\vspace{-.5em}\vspace{2pt}}
\end{center}\eightpt\list
 {[\arabic{enumi}]}{\settowidth\labelwidth{[#1]}\leftmargin\labelwidth
 \advance\leftmargin\labelsep
 \usecounter{enumi}}
 \def\newblock{\hskip .11em plus .33em minus .07em}
 \sloppy\clubpenalty4000\widowpenalty4000
 \sfcode`\.=1000\relax}

\title{RWTH ASR Systems for LibriSpeech: Hybrid vs Attention\\ - w/o Data Augmentation}
\name{Christoph L\"uscher$^1$, Eugen Beck$^{1,2}$, Kazuki Irie$^1$, Markus Kitza$^1$, Wilfried Michel$^{1,2}$, Albert Zeyer$^{1,2}$,
Ralf Schl\"uter$^1$, Hermann Ney$^{1,2}$}
\address{
$^1$Human Language Technology and Pattern Recognition,
Computer Science Department, \\
RWTH Aachen University, 52074 Aachen, Germany \\
$^2$AppTek GmbH, 52062 Aachen, Germany}%
\email{\{luescher, beck, irie, kitza, michel, zeyer, schlueter, ney\}@cs.rwth-aachen.de}

\begin{document}

\maketitle
\begin{abstract}
We present state-of-the-art automatic speech recognition (ASR) systems employing a standard hybrid DNN/HMM architecture compared to an attention-based encoder-decoder design for the LibriSpeech task.
Detailed descriptions of the system development, including model design, pretraining schemes, training schedules, and optimization approaches are provided for both system architectures.
Both hybrid DNN/HMM and attention-based systems employ bi-directional LSTMs for acoustic modeling/encoding.
For language modeling, we employ both LSTM and Transformer based architectures.
All our systems are built using RWTH’s open-source toolkits RASR and RETURNN.
To the best knowledge of the authors, the results obtained when training on the full LibriSpeech training set, are the best published currently, both for the hybrid DNN/HMM and the attention-based systems.
Our single hybrid system even outperforms previous results obtained from combining eight single systems.
Our comparison shows that on the LibriSpeech 960h task, the hybrid DNN/HMM system outperforms the attention-based system by 15\% relative on the clean and 40\% relative on the other test sets in terms of word error rate.
Moreover, experiments on a reduced 100h-subset of the LibriSpeech training corpus even show a more pronounced margin between the hybrid DNN/HMM and attention-based architectures.

\end{abstract}

\section{Introduction}
\label{sec:intro}

Over the last years, \ac{asr} systems have improved significantly.
Especially the rise of deep \acp{nn} has accelerated this development immensely \cite{hinton2012:deep}.
Convolutional \acp{nn} and recurrent \acp{nn} are the state-of-the-art architectures for most \ac{asr} tasks.
State-of-the-art systems are largely based on the hybrid deep neural network (DNN) based standard architectures.
However, the general progress in deep learning/machine learning also triggered a diversification of ASR architectures into a series of so-called end-to-end approaches.
Most notably, this includes the attention-based encoder-decoder architecture, for which good performance has been reported on a number of tasks, including the LibriSpeech task.

The LibriSpeech task comprises English read speech data based on the LibriVox project \cite{panayotov2015:librispeech}.
Previous results on LibriSpeech using hybrid models are presented in \cite{han2017:capio,kanda2018:lattice}.
While \cite{han2017:capio} uses a \ac{gmmhmm} \acuse{gmm} \acuse{hmm} as the basis for their system, further training is conducted with a hybrid \acs{dnn}/\ac{hmm} with a densely connected topology.
The densely connected \acp{nn} in \cite{han2017:capio} are composed of different types of \ac{nn} layers: convolutional \acp{nn}, \acl*{tdnn} and bi-directional \acp{lstm}.
\Ac{lfmmi} is applied during training.
A recurrent \ac{nn} \ac{lm} is used for rescoring.
The final best result in \cite{han2017:capio} is achieved with a system combination of eight systems.
In \cite{kanda2018:lattice}, a lattice-free \ac{smbr} training method is used.

End-to-end results on LibriSpeech were presented in
\cite{zeyer2018:asr-attention,zeyer2018:attanalysis,sabour2018ocd,zeghidour2018:fully,irie2019}.
The end-to-end approach in \cite{zeghidour2018:fully} uses the raw waveform and a convolutional \ac{nn} acoustic model with gated linear units.
An end-to-end attention-based encoder-decoder approach with a pretraining scheme is presented in \cite{zeyer2018:asr-attention,zeyer2018:attanalysis}.
In \cite{sabour2018ocd} a training procedure based on edit distance for sequence to sequence model optimization is presented.
An exploration of target units (phoneme, grapheme and word-piece) in relation to training size was performed in \cite{irie2019}.
A data augmentation method called SpecAugment was presented in \cite{park2019specaugment}.
So far, while end-to-end approaches show competitive performance, they are outperformed by hybrid approaches.
We compare the conventional hybrid \acs{dnn}/\acs{hmm} approach on phone level
to the encoder-decoder-attention model which directly operates on the word or sub-word level
and is thus often referred to as an end-to-end model.
In addition, we use word-level and subword-level neural language models to further improve the performance of both systems.
We describe the development of our hybrid system and show which factors were especially important for the performance of the system.
To the best of our knowledge, the results obtained on the LibriSpeech task reflect state-of-the-art performance for both hybrid and attention-based modeling, with a clear margin still for hybrid DNN/HMM modeling when no data augmentation scheme is applied.

\section{Hybrid system}
\label{sec:gmmhmm}

\subsection{Acoustic model}
\label{sec:am}

\subsubsection{\Ac{gmmhmm} system}

We use 16-dim \ac{mfcc} adding first and second order derivatives, and additionally energy features as input for the \ac{gmmhmm} system.
The transition probabilities are set manually and applied to all \acp{hmm}.

The first step is linear time alignment where the features are uniformly distributed over the audio.
We iterate the repetition of the parameter estimation based on the linear time alignment five times.
Afterwards, we perform a non-linear time alignment to improve the alignment.
Afterwards we perform parameter estimation.
Initially, this process was iterated 10 times.
Increasing the number of iterations showed constant improvement.
Therefore we continued adding training iterations until \ac{wer} convergence.
Training of a state-tied triphone \ac{gmmhmm} model is the following step.
The states are tied using a phonetic \ac{cart}.
We experimented with different numbers of \ac{cart} labels ranging from $9k-20k$ plus silence.
All state-tied triphones use three \ac{hmm} states.
We switch the input features from 16-dim. \ac{mfcc} with derivatives to a context window of \ac{mfcc} features resulting in a 144-dim. feature vector on which \ac{lda} is performed.
The \ac{lda} output has a dimension of 48.
After training the state-tied triphone \ac{gmmhmm} model, \ac{vtln} is applied, followed by \ac{sat} with \acl*{cmllr} to adapt the Gaussian mixture model parameters to a speaker.
After adapting the parameters, a realignment is performed.

\subsubsection{Hybrid \ac{dnn}/\ac{hmm} system}

The \ac{nn} acoustic model architecture is a bi-directional \ac{lstm} \cite{hochreiter1997:lstm,graves2013hybrid}.
This architecture achieves good performance in acoustic modelling \cite{graves2013hybrid,zeyer17:lstm}.
For the hybrid \ac{dnn}/\ac{hmm} system we extract several different features: 16-dim \ac{mfcc} with derivatives and energy, 48-dim. features from the triphone system and Gammatone filters \cite{schlueter:icassp07} with 25, 50 or 100-dim.
The extracted 50-dim. Gammatone filters had the best performance.
All features are used as input into the bi-directional \ac{lstm} along with the generated alignments from the \ac{gmmhmm} system.
We continue to use \ac{cart} labels for the state-tied phones.
The same range of \ac{cart} labels was used for experimentation.
The network topology consists of six bi-directional \ac{lstm} layers with 1000 units for backward and forward direction each.
We experimented with smaller bi-directional \ac{lstm} sizes (number of layers and number of units per layer) but found them to be worse in performance.
The output layer is comprised of a softmax layer with output units corresponding to the number of the \ac{cart} labels.
Frame-wise cross-entropy loss criterion and Adam optimization with Nesterov momentum (Nadam) are used for the mini-batch training of the network \cite{kingma2014:adam,dozat2016:incorporating}.
Newbob learning rate scheduling \cite{zeyer17:lstm} is applied to control the learning rate reduction with a learning rate decay rate of $0.9$.
$L_2$ regularization was used to prevent overfitting.
The $L_2$ hyperparameter was set to $0.01$.
Further regularization is done with dropout \cite{srivastava2014:dropout}.
We experimented with dropout in the range of $5$\% -- $30$\% and found a dropout of $10$\% to work best for us.
Gradient noise \cite{neelakantan2015:adding} with a variance of $0.1$ was employed.
We experimented with different learning rates and batch sizes in various combinations.
So far a batch size of $20k$ and a learning rate of $0.008$ have shown the best performance.
Additionally, learning rate warm up proved to be helpful.
We start with a learning rate of $0.003$ and increase the learning rate to $0.008$ over the first ten subepochs.
A subepoch is 1/40th of the training data.
The training data is seen $12.5$ times.
During decoding the \ac{lm} scale is an important hyperparameter which will effect the \ac{wer} directly.
We found a scale between $11$--$13$ worked best for us.

\subsubsection{Sequence discriminative training}
\label{sec:seqdisc}
Sequence discriminative training is performed using a lattice-based version of the state-level minimum Bayes risk (sMBR) criterion \cite{mbr}.

The hybrid \ac{dnn}/\ac{hmm} model is used to generate lattices for all of the training data.
The training is then continued from the hybrid \ac{dnn}/\ac{hmm} model with a lower learning rate.

We use cross-entropy smoothing with a smoothing factor of $0.1$ and early stopping to prevent overfitting.

\subsection{Language model}
\label{sec:lm}
We report performance of hybrid systems using both a 4-gram count based language
model \cite{kneser1995improved} and an LSTM language model \cite{sundermeyer12:lstm} in the \textit{first pass decoding} \cite{beck2019:lstmlm1pass}.
We use the 4-gram count model officially distributed with the LibriSpeech dataset \cite{panayotov2015:librispeech}.
For the LSTM language model, we train our own model using our toolkit RETURNN \cite{zeyer2018:returnn}.
Two training datasets are available for language modeling: 800M-word text only data and
960h of audio transcriptions which corresponds to 10M-word text data. These two sets are merged to form one
training dataset for language model training.
Our LSTM language model has two recurrent layers with 4096 LSTM nodes in each layer, an input projection
layer of size 128, and a output softmax layer over the full 200k vocabulary.
We train the model using the stochastic gradient descent with gradient norm clipping and Newbob
learning rate scheduling.

In addition, we carry out \textit{rescoring of lattices} generated by the LSTM language model using a Transformer \cite{transfo} language model.
Our Transformer model has 96 layers with the self-attention total dimension of 512 using 8 heads and the inner feed-forward dimension of 2048 in each layer, which gave the best development perplexity in our preliminary experiments \cite{irie:is19}.
We use push-forward algorithm \cite{sundermeyer2014:rescoring} with recombination pruning of
order 9.
We linearly interpolate the two models with interpolation weights optimized on the development perplexity.
We found 0.71 to be the optimal weight on the Transformer model which gave the development perplexity of 52.3,
while the LSTM and Transformer models have the individual development perplexity of 60.2 and 53.7 respectively.
\begin{table*}[!t]
\centering
\caption{\ac{gmmhmm} and hybrid \ac{dnn}/\ac{hmm} results on LibriSpeech with 12k \ac{cart} labels and evaluated with the official 4-gram \ac{lm}.}
\label{tab:evol}
\begin{tabular}{|c|c|c|c|c|c|c|c|c|}
\hline
\multirow{3}{*}{phoneme context} & \multirow{3}{*}{acoustic model} & \multirow{3}{*}{\ac{vtln}} & \multirow{3}{*}{\ac{sat}} & \multirow{3}{*}{sMBR} & \multicolumn{4}{c|}{\acs*{wer} {[}\%{]}} \\ \cline{6-9}
&&&&                                                                                      & \multicolumn{2}{c|}{dev}           & \multicolumn{2}{c|}{test} \\ \cline{6-9}
&&&&                                                                                      & clean & \multicolumn{1}{c|}{other} & clean & \multicolumn{1}{c|}{other} \\ \hline \hline
mono&\multirow{5}{*}{GMM}&\multirow{2}{*}{no} & \multirow{3}{*}{no} & \multirow{6}{*}{no} & 24.3 & 52.6  & 24.1 & 56.1 \\ \cline{1-1}\cline{6-9}
\multirow{6}{*}{tri}    &&                    &                     &                     & 12.1 & 34.5  & 12.9 & 36.9 \\ \cline{3-3}\cline{6-9}
&                        & yes                &                     &                     & 12.0 & 35.1  & 11.2 & 36.4  \\ \cline{3-4}\cline{6-9}
&                        & no                 & \multirow{4}{*}{yes}&                     &  8.0 & 21.9  &  8.6 & 22.9  \\ \cline{3-3}\cline{6-9}
&                        &\multirow{3}{*}{yes}&                     &                     &  7.6 & 22.0  &  8.4 & 23.1  \\ \cline{2-2}\cline{6-9}
&  \multirow{2}{*}{LSTM} &                    &                     &                     &  4.0 &  9.6  &  4.4 & 10.0  \\ \cline{5-9}
&                        &                    &                     & yes                 &  3.4 &  8.3  &  3.8 &  8.8  \\ \hline
\end{tabular}
\end{table*}

\section{Encoder-Decoder-Attention system}
\label{sec:att}

The \emph{encoder-decoder framework with attention} has initially been introduced for machine translation
where it dominates the field now
\cite{bahdanau2014nmt,luong2015attentionmt,chen2018rnmtplus}.
Recent investigations have shown promising results by applying the same approach for
speech recognition \cite{
chan2016las, %
doetsch2016:bidir-dec-att, %
chiu2017sotaasratt, %
battenbergAsru17,
sabour2018ocd,
zeyer2018:asr-attention}.
Among end-to-end approaches for \ac{asr}, the attention model seems to perform best
\cite{zeyer2018:attanalysis}.
Our model operates on sub-word units via byte-pair encoding \cite{sennrich16bpe}.
As input 40-dim MFCC feature vectors are used.
Our presented results outperform the best LibriSpeech attention system presented in \cite{zeyer2018:attanalysis}.
Compared to the system in \cite{zeyer2018:attanalysis} we use an extended pretraining variant where we not only grow the encoder depth but also grow the hidden dimension of the LSTMs.
Specifically, we start with 2 layers in the encoder of dimension 512
and increase to 6 layers with dimension 1024.
Additionally, we train the first pretrain construction step first without dropout.
We improved upon that model by tuning the curriculum learning schedule slightly,
i.e.\ we have these 4 steps with different portions of the dataset:
\begin{enumerate}
\item from 25\% of the whole data, take only train-clean,
and filter randomly such that the max mean number of characters in the transcriptions of each sequence
is 50,
\item from the next 25\% of the whole data, take only train-clean,
and filter randomly such that the max mean number of characters is 75,
\item from the next 50\% of the whole data, take only train-clean,
\item from now on, take everything.
\end{enumerate}
Also, in the pretraining, we repeat the first step once more, with 2 layers of dimension 512, without dropout.
The next improvement came from just training longer,
i.e.\ we trained with our learning rate scheduling until it converged,
then took the best model, and continued training with a reset learning rate scheduling.
We repeated this twice.
In the first iteration, we went over the whole data 12.5 times, then another 6.6 times and finally another 8.3 times,
i.e. in total 27.4 times.%

To further enhance end-to-end system's performance, we train \ac{bpe}-level language models and
apply them to the system by shallow fusion\cite{gulcehre+al-2016-monolingual, toshniwal2018comparison}.
We report the performance of LSTM based and Transformer based language models separately.
Our LSTM model has 4 recurrent layers with 2048 LSTM nodes.
We use a 24-layer Transformer model with 8-head self-attention and feed-forward dimensionality of 1024 and 4096 respectively,
which we obtained in \cite{irie:is19}.
We select the language model checkpoints for the recognition experiments based on the development perplexity.
For shallow fusion, we apply a single weight on the language model score (the weight on the score of the attention model is 1)
and we use a beam size of 64 as well as an end-of-sentence penalty \cite{hannun2019sequence}.
We optimize the weights separately on the dev-clean and dev-other sets, then respectively apply them to the test-clean and test-other sets.
We found optimal weights to be similar for both models; 0.5 and 0.56 for the LSTM language model,
and 0.52 and 0.54 for the Transformer model, respectively on the clean and other sets.

\section{Experimental setup}
\label{sec:setup}

The two systems, a hybrid-\ac{dnn} and an attention-based encoder-decoder are both trained on the 960h training data from the LibriSpeech corpus.
For comparison, also a 100h subset is used.
Unless specified otherwise, the training was performed using the full training set of 960h.
The data is in English but the content ranges from different time periods and different English speaking countries.
Having the consequence of different English styles being within the corpus.

The hybrid model was trained and decoded with RASR \cite{wiesler2014:rasr} and RETURNN \cite{zeyer2018:returnn,doetsch2017:returnn}.
The monophone and triphone system to generate the alignments was built in RASR while the \ac{nn} model was trained in RETURNN.
The decoding process was setup in RASR.
Our encoder-decoder-attention model was trained and decoded using RETURNN \cite{zeyer2018:returnn}.
Both toolkits are open-source.
All the config files used for training and recognition of all our results are publicly available online \cite{returnnexperiments2019}.

We evaluate the models on the dev and test sets provided with the LibriSpeech corpus: \emph{dev-clean}, \emph{dev-other}, \emph{test-clean} and \emph{test-other}.
The difference between \emph{clean} and \emph{other} is the quality of the audio and its corresponding transcription.
The \emph{clean} quality is higher than the \emph{other}.

\section{Experimental results}
\label{sec:results}

The development stages of our acoustic model are shown in \Cref{tab:evol}.
We start the training of the \ac{gmmhmm} model from scratch using linear alignments.
Afterwards we utilize non-linear alignments.
To further improve the \ac{gmmhmm} model we introduce triphones.
Adding \ac{vtln} on top of the triphone system only shows improvements on clean but degradation on other.
However adding \ac{sat} to the triphone system improves the \ac{wer}.
Combining \ac{vtln} and \ac{sat} gives mixed \ac{wer}: \emph{clean} improves, \emph{other} degrades.
Introducing an hybrid \ac{dnn}/\ac{hmm} improves the system \ac{wer} results.
Continuing with sequence discriminative training improves the performance even further.

\begin{table}[!ht]
\centering
\caption{Hybrid \ac{dnn}/\ac{hmm} results on LibriSpeech with different numbers of \ac{cart} labels. For all systems the official 4-gram word \ac{lm} is used.}
\label{tab:num}
\begin{tabular}{|c|c|c|c|c|}
\hline
\multirow{3}{*}{\# of \ac{cart} labels} & \multicolumn{4}{c|}{\acs*{wer} {[}\%{]}} \\ \cline{2-5}
                       & \multicolumn{2}{c|}{dev}           & \multicolumn{2}{c|}{test} \\ \cline{2-5}
                       & clean & other & clean & other \\ \hline \hline
9001                   &  6.2  & 14.9  &  5.8  & 15.9  \\ \hline
12001                  &  4.0  &  9.6  &  4.4  & 10.0  \\ \hline
20001                  &  4.9  & 11.3  &  5.4  & 12.3  \\ \hline
\end{tabular}
\end{table}

We evaluated the influence of the number of \ac{cart} labels with the hybrid \ac{dnn}/\ac{hmm} model and the official 4-gram \ac{lm} (\Cref{tab:num}).
9k \ac{cart} labels show the worst performance.
In contrast, 20k \ac{cart} labels shows improved performance.
But the best performance was shown by 12k \ac{cart} labels.

\begin{table}[!ht]
\setlength{\tabcolsep}{3pt}
\centering
\caption{Comparison between hybrid \ac{dnn}/\ac{hmm} and encoder-decoder-attention model results on LibriSpeech with different training corpus sizes.
train-clean-100 is a official subset of the training corpus.
train-960 is the complete training corpus.
(Clustered) context-dependent phones (CDp) are utilized for the hybrid model,
and sub-word units for the attention model.}
\label{tab:size}
\begin{tabular}{|l|c|c|c|c|c|c|}
\hline
\multirow{3}{*}{training set} & \multirow{3}{*}{model} & \multirow{3}{*}{LM} & \multicolumn{4}{c|}{\acs*{wer} {[}\%{]}} \\ \cline{4-7}
                &         &  & \multicolumn{2}{c|}{dev} & \multicolumn{2}{c|}{test} \\ \cline{4-7}
                &         &  & clean & other & clean & other \\ \hline \hline
\multirow{2}{*}{train-clean-100} & hybrid    & 4-gram &  5.0  & 19.5  & 5.8  & 18.6   \\ \cline{2-7}
                & attention & none &  14.7     &   38.5    &  14.7    &  40.8      \\ \hline
\multirow{2}{*}{train-960}      & hybrid   & 4-gram  &  4.0  &  9.6  & 4.4  & 10.0   \\ \cline{2-7}
                & attention & none &   4.7   &   14.3   &  4.8  &  15.4   \\ \hline
\end{tabular}
\end{table}

\begin{table*}[!t]
\centering
\setlength{\tabcolsep}{0.55em}
\caption{
The \ac{wer} results from our most interesting models and important results from other papers on LibriSpeech 960 h. CDp are (clustered) context-dependent phones. \ac{bpe} are sub-word units. 4-gr \ac{lm} is the official 4-gram word \ac{lm}. GCNN are gated convolutional \ac{nn}. RNN are recurrent \ac{nn}.}
\label{tab:wer}
\begin{tabular}{|l|c|c|c|c|c|c|c|c|}
\hline
\multirow{3}{*}{paper} & \multirow{3}{*}{model} & \multicolumn{2}{c|}{label unit} & \multirow{3}{*}{LM} & \multicolumn{4}{c|}{\acs*{wer} {[}\%{]}} \\ \cline{3-4}\cline{6-9}
&                      & \multirow{2}{*}{AM}  & \multirow{2}{*}{LM} &    & \multicolumn{2}{c|}{dev} & \multicolumn{2}{c|}{test} \\ \cline{6-9}
&                      &          &        &    & clean & other & clean & other \\ \hline \hline
\multirow{2}{*}{Han et al. \cite{han2017:capio}}& hybrid, seq. disc., single & \multirow{2}{*}{CDp} & \multirow{2}{*}{word} & \multirow{2}*{RNN} & 3.0 &  8.8 & 3.6 &  8.7 \\ \cline{2-2}\cline{6-9}
& hybrid, seq. disc., ensemble                                               &                      &                       &                    & 2.6 &  7.6 & 3.2 &  7.6 \\ \hline
Zeghidour et al. \cite{zeghidour2018:fully} & end-to-end GCNN & chars           & words                 & GCNN               & 3.2 & 10.1 & 3.4 & 11.2 \\ \hline
Irie et al. \cite{irie2019}&\multirow{5}{*}{end-to-end attention}&\multicolumn{2}{c|}{Word Piece Model}&\multirow{2}{*}{\ac{lstm}}& 3.3 & 10.3 & 3.6 & 10.3 \\ \cline{1-1}\cline{3-4}\cline{6-9}
Zeyer et al. \cite{zeyer2018:asr-attention}&&\multicolumn{2}{c|}{\multirow{4}{*}{\ac{bpe}}}&                              & 3.5 & 11.5  & 3.8 & 12.8 \\ \cline{1-1}  \cline{5-9}
\multirow{7}{*}{this work} &                & \multicolumn{2}{c|}{} & None        & 4.3 & 12.9  & 4.4 & 13.5 \\ \cline{5-9}
&                                           & \multicolumn{2}{c|}{}                        & \ac{lstm}   & 2.9 &  8.9  & 3.2 & 9.9 \\ \cline{5-9}
&                                           & \multicolumn{2}{c|}{}                        & Transformer & \textbf{2.6} & \textbf{8.4} & \textbf{2.8} & \textbf{9.3} \\ \cline{2-9}
& hybrid                                    & \multirow{4}{*}{CDp}      & \multirow{4}{*}{word} & \multirow{2}{*}{4-gr} &  4.0  &  9.6  &  4.4  & 10.0  \\ \cline{2-2}\cline{6-9}
& \multirow{3}{*}{hybrid, seq. disc.}       &                           &                       &                       &  3.4  &  8.3  &  3.8  &  8.8  \\ \cline{5-9}
&                                           &                           &                       & +  \ac{lstm} & \textbf{2.2} & \textbf{5.1} & \textbf{2.6} & \textbf{5.5} \\ \cline{5-9}
&                                           &                           &                       & Transformer resc. & \textbf{1.9}  & \textbf{4.5} & \textbf{2.3} & \textbf{5.0} \\ \hline \hline
Park et. al.\cite{park2019specaugment}&  end-to-end attention/SpecAugment &\multicolumn{2}{c|}{Word Piece Model} &      LSTM              & - & - & 2.5 & 5.8 \\ \hline
\end{tabular}
\end{table*}

We compare the hybrid model with the encoder-decoder-attention model.
We trained both models on the \emph{train-clean-100} training subset and on the \emph{train-960} complete training set.
These are not the best models but utilize a baseline model for both approaches.
The hybrid \ac{dnn}/\ac{hmm} model outperforms the encoder-decoder-attention model constantly.
But the difference in performance shrinks substantially with the much larger training set.

Our encoder-decoder-attention model in combination with a Transformer \ac{lm} gives a \ac{wer} of $3.2$\% on \emph{test-clean} and $9.9$\% on \emph{test-other} (\Cref{tab:wer}).
Evaluating our sequence discriminativly trained acoustic model with our \ac{lstm} \ac{lm} results in a \ac{wer} of $2.6$\% on \emph{test-clean} and $5.5$\% on \emph{test-other}.
Rescoring with a Transformer language model further improves the performance of our hybrid \ac{dnn}/\ac{hmm} system
resulting in a \ac{wer} of are $2.3$\% on \emph{test-clean} and $5.0$\% on \emph{test-other}.
The previous best hybrid system was presented in \cite{han2017:capio} while the best end-to-end system without data augmentation was presented in \cite{zeghidour2018:fully,irie2019} (\Cref{tab:wer}).
Additionally we present the best end-to-end system with data augmentation \cite{park2019specaugment}.
Our best encoder-decoder-attention model improves the state-of-the-art for end-to-end models without data augmentation by $17.6$\% relative \ac{wer} on \emph{test-clean} and by $3.9$\% relative \ac{wer} on \emph{test-other}.
Our best hybrid \ac{dnn}/\ac{hmm} system without Transformer \ac{lm} rescoring improves the state-of-the-art by $18.8$\% relative \ac{wer} on \emph{test-clean} and by $27.6$\% relative \ac{wer} on \emph{test-other}.
If we add rescoring with a Transformer \ac{lm} we improve further by $11.5$\% relative \ac{wer} on \emph{test-clean} and by $9.1$\% relative \ac{wer} on \emph{test-other}.
In comparison, the hybrid \ac{dnn}/\ac{hmm} system still outperforms the encoder-decoder-attention system by over $15$\% relative \ac{wer} on \emph{test-clean} and by over $40$\% relative \ac{wer} on \emph{test-other}.
Our best hybrid model even outperforms the end-to-end attention model with SpecAugment \cite{park2019specaugment} by $8$\% relative \ac{wer} on \emph{test-clean} and by $13.8$\% relative \ac{wer} on \emph{test-other}.
These results reflect the state-of-the-art performance for both hybrid and attention-based models on LibriSpeech, to the best of the authors' knowledge.

\acp{wer} become very small, especially for \emph{dev-clean} and \emph{test-clean}.
When analyzing the errors, it is noticeable that some of the errors would not be recognized as primary errors by a human.
These can be categorized as, for example: word contractions or American vs British English spelling.
Examples of such errors are: I am $\leftrightarrow$ I'm, tyrannise $\leftrightarrow$ tyrannize, color $\leftrightarrow$ colour, oh $\leftrightarrow$ o.
So far we have not employed a normalization strategy for these errors.

\section{Conclusions}

In this paper we presented two \ac{asr} systems for the LibriSpeech task.
One System was a hybrid \ac{dnn}/\ac{hmm} system based on a \ac{gmmhmm} system, the other system was an attention-based encoder-decoder system.\par
We described how we built the systems and described how to incrementally improve the systems to get competitive results.
For the hybrid \ac{dnn}/\ac{hmm} system a large \ac{nn} acoustic model, the sequence discriminative training and the employment of an \ac{lstm} \ac{lm} was important for the good performance.
The encoder-decoder-attention approach utilized an extended pretraining variant and a tuned curriculum learning schedule.
This enabled the model to achieve competitive results in comparison to other end-to-end approaches.

The presented encoder-decoder-attention system showed state-of-the-art performance on the LibriSpeech 960h task in comparison with end-to-end systems without data augmentation.
But our comparison shows that on the LibriSpeech 960h task, the hybrid DNN/HMM system outperforms the attention-based system by $15$\% relative on the clean and $40$\% relative on the other test sets.
Our hybrid system even outperforms previous results presented in the literature.
Moreover, experiments on a reduced 100h-subset of the LibriSpeech training corpus even show a more pronounced margin between the hybrid DNN/HMM and attention-based architectures.
To the best knowledge of the authors, the results obtained when training on the full LibriSpeech training set, are the best published currently, both for the hybrid DNN/HMM and the attention-based systems presented in this work.

\acknowledgements

\begin{footnotesize}

This work has received funding from the European Research Council (ERC)
under the European Union's Horizon 2020 research and innovation programme
(grant agreement No 694537, project "SEQCLAS")
and from a Google Focused Award.
The work reflects only the authors' views and
none of the funding parties is responsible for any
use that may be made of the information it contains.

Experiments were partially performed with computing resources granted by RWTH Aachen University under project nova0003.

We thank Wei Zhou for help with generating lattices.

\end{footnotesize}

\bibliographystyle{IEEEtran}

\bibliography{mybib}

\end{document}